\documentclass{article} % For LaTeX2e
\usepackage{nips13submit_e,times}
\usepackage{hyperref}
\usepackage{url}

\usepackage{epsfig}
\usepackage{picinpar}
\usepackage{graphicx}
\usepackage{amsmath}
\usepackage{amsthm}
\usepackage{amssymb,bm}
\usepackage[ruled,vlined]{algorithm2e}
\usepackage[titletoc]{appendix}
\usepackage{float,caption,multirow}

\newcommand{\diag}{\operatornamewithlimits{Diag}}
\newcommand{\sgn}{\operatornamewithlimits{sgn}}
\newcommand{\argmin}{\operatornamewithlimits{argmin}}
\newcommand{\subto}{\operatornamewithlimits{s.t.}}

\newcommand{\defeq}{\mathrel{\mathop:}=}

\newcommand{\mc}{\mathcal}
\newcommand{\bs}{\mathbf}
\newcommand{\mb}{\mathbb}

\newtheorem{definition}{Definition}
\newtheorem{proposition}{Proposition}
\newtheorem{theorem}{Theorem}
\newtheorem{lemma}{Lemma}

\title{Exclusivity Regularized Machine}

\author{
Xiaojie Guo\\
State Key Laboratory Of Information Security\\
Institute of Information Engineering, Chinese Academy of Sciences\\
\texttt{xj.max.guo@gmail.com} \\
}

\nipsfinalcopy % Uncomment for camera-ready version

\begin{document}

\maketitle

\begin{abstract}
It has been recognized that the diversity of base learners is of utmost importance to a good ensemble. This paper defines a novel measurement of diversity, termed as \emph{exclusivity}. With the designed exclusivity, we further propose an ensemble model, namely \emph{Exclusivity Regularized Machine} (ERM), to jointly suppress the training error of ensemble and enhance the diversity between bases. Moreover, an Augmented Lagrange Multiplier based algorithm is customized to effectively and efficiently seek the optimal solution of ERM. Theoretical analysis on convergence and global optimality of the proposed algorithm, as well as experiments are provided to reveal the efficacy of our method and show its superiority over state-of-the-art alternatives in terms of accuracy and efficiency.  
\end{abstract}

\section{Introduction}
%$\mc{H}$ $\mc{D}$ $\mc{I}$
Classification is a major task in the fields of machine learning and pattern recognition. In binary classification, a hypothesis is constructed from a feasible hypothesis space based on the training set $\{(\bs{x}_i, y_i)\}_{i=1}^N$, where $\{\bs{x}_i\}_{i=1}^N$ is a set of data points with $\bs{x}_i\in\mb{R}^d$ sampled i.i.d. under a distribution from an input subspace, and $\{y_i\}_{i=1}^N$ with $y_i\in\{-1,+1\}$ are their corresponding labels. The obtained hypothesis, also known as classifier, is ``good" when it is able to generalize well the ``knowledge" learned from the training data to unseen instances. Multiple-class cases can be analogously accomplished by a group of binary classifiers \cite{EnsembleM}. 

Arguably, among existing classifiers, \emph{Support Vector Machine} (SVM) \cite{SVM}\cite{PSVM} is the most popular one due to its promising performance. In general, the primal SVM can be modeled as follows:
\begin{equation}
\argmin_{\{\bs{w}, b\}} \Psi(\bs{w})+\lambda\sum_{i=1}^N f(y_i,\phi(\bs{x}_i)^T\bs{w}+b),
\label{eq:svmg}
\end{equation}
where $f(\cdot)$ is a penalty function, $\Psi(\bs{w})$ performs as a regularizer on the learner $\bs{w}$ and $b$ is the bias. The function $\phi(\cdot)$ is to map $\bs{x}_i$ from the original $D$-dimensional feature space to a new $M$-dimensional one. Moreover, $\lambda$ is a non-negative coefficient that provides a trade-off between the loss term and the regularizer. 
If SVM adopts the hinge loss as penalty, the above \eqref{eq:svmg}  turns out to be: 
\begin{equation}
\argmin_{\{\bs{w}, b\}} \Psi(\bs{w})+\lambda\sum_{i=1}^N \big(1-(\phi(\bs{x}_i)^T\bs{w}+b)y_i\big)_{+}^p,
\label{eq:svmh}
\end{equation}
where the operator $(u)_{+}\defeq\max(u, 0)$ keeps the input scalar $u$ unchanged if $u$ is non-negative, returns zero otherwise, the extension of which to vectors and matrices is simply applied element-wise. Furthermore, $p$ is a constant typically in the range $[1, 2]$ for being meaningful. In practice, $p$ is often selected to be either $1$ or $2$ for ease of computation, which correspond to $\ell^1$-norm and $\ell^2$-norm loss primal SVMs, respectively. As for the regularization term, $\Psi(\bs{w})\defeq \frac{1}{2}\|\bs{w}\|_2^2$ ($\ell^2$ regularizer) and $\Psi(\bs{w})\defeq\|\bs{w}\|_1$ ($\ell^1$ regularizer) are two classical options. %The attractiveness of primal SVM is basically from that the primal objective function is guaranteed to be continuously decreased .

As has been well recognized, a combination of various classifiers can improve predictions. Ensemble approaches, with Boosting \cite{Adaboost} and Bagging \cite{Bagging} as representatives, make use of this recognition and achieve strong generalization performance. The generalization error of ensemble mainly depends on two factors, formally expressed as $E = \bar{{E}}-\bar{{A}}$, where $E$ is the mean-square error of the ensemble, $\bar{E}$ represents the average mean-square error of component learners and $\bar{A}$ stands for the average difference (diversity) between the ensemble and the components.  \emph{Error-Ambiguity decomposition} \cite{EAD}, \emph{Bias-Variance-Covariance decomposition} \cite{BVC} and \emph{Strength-Correlation decomposition} \cite{SCD} all confirm the above principle. This indicates that jointly minimizing the training error and maximizing the diversity of base learners is key to the ensemble performance. Considering the popularity of SVM and the potential of ensemble, it would be interesting and beneficial to equip SVM with ensemble thoughts. 
% which train and integrate a set of component learns

This work concentrates on how to train a set of component SVMs and integrate them as an ensemble. More concretely, the contribution of this paper can be summarized as follows: 1) we define a new measurement, namely (relaxed) exclusivity, to manage the diversity between base learners, 2) 2e propose a novel ensemble, called Exclusivity Regularized Machine (ERM), which concerns the training error and the diversity of components simultaneously, and 3) we design an Augmented Lagrange Multiplier based algorithm to efficiently seek the solution of ERM, the global optimality of which is theoretically guaranteed.
%\begin{itemize}
%\item{We define a new measurement, namely (relaxed) exclusivity, to manage the diversity between base learners.}
%\item{We propose a novel ensemble, called Exclusivity Regularized Machine (ERM), which concerns the training error and the diversity of components simultaneously.}
%\item{We design an Augmented Lagrange Multiplier based algorithm to efficiently seek the solution of ERM, the global optimality of which is theoretically guaranteed.}
%%\item{To reveal the efficacy and the superior performance of the proposed ERM in comparison with state-of-the-arts, experiments on several benchmarks are conducted.}
%\end{itemize}

\section{Exclusivity Regularized Machine}

\subsection{Definition and Formulation}
\label{sec:d&f}
It is natural to extend the traditional primal SVM \eqref{eq:svmh} to the following ensemble version with $C$ components as:
\begin{equation}
\argmin_{\{\bs{W}, \bs{b}\}} \Psi(\bs{W})+\lambda\sum_{c=1}^C\sum_{i=1}^N \big(1-(\phi(\bs{x}_i)^T\bs{w}_c+b_c)y_i\big)_{+}^p,
\label{eq:svmm}
\end{equation}
where $\bs{W}\in\mb{R}^{M\times C}\defeq[\bs{w}_1,\bs{w}_2,...,\bs{w}_C]$ and $\bs{b}\in\mb{R}^{C} \defeq [b_1,b_2,..,b_C]^T$. Suppose we simply impose $\frac{1}{2}\|\bs{W}\|_F^2$ or $\|\bs{W}\|_1$ on $\bs{W}$, all the components (and the ensemble) will have no difference with the hypothesis directly calculated from \eqref{eq:svmh} using the same type of regularizer. From this view, $\Psi(\bs{W})$ is critical to achieve the diversity.\footnote{We note that splitting the training data into $C$ sub-sets and training $C$ classifiers separately on the sub-sets would lead to some difference between the components. But, this strategy is not so reliable since if the training data are sufficiently and i.i.d. sampled under a distribution, the difference would be very trivial.}

Prior to introducing our designed regularizer, we first focus on the concept of diversity. Although the diversity has no formal definition so far, the thing in common among studied measurements is that the diversity enforced in a pairwise form between members strikes a good balance between complexity and effectiveness. The evidence includes Q-statistics measure \cite{Q-S}, correlation coefficient measure \cite{Q-S}, disagreement measure \cite{Dis-M}, double-fault measure \cite{D-F}, $k$-statistic measure \cite{k-S} and mutual angular measure \cite{DRM,DIM1,DIM2}. These measures somehow enhance the diversity, however, most of them are heuristic. One exception is Diversity Regularized Machine \cite{DRM}, which attempts to seek the globally-optimal solution. Unfortunately, it often fails because the condition required for the global optimality, say $\|\bs{w}_c\|_2=1$ for all $c$, is not always satisfied. Further, Li \textit{et al.} proposed a pruning strategy to improve the performance of DRM \cite{DREP}. But, DRM requires too much time to converge, which limits its applicability. In this work, we define a new measure of diversity, \textit{i.e.} (relaxed) exclusivity, as below.   %the angle measure of diversity used by DRM, say $\diver(\bs{u},\bs{v})\defeq1-\frac{\bs{u}^T\bs{v}}{\|\bs{u}\|_2\cdot\|\bs{v}\|_2}$, is a bit weak, and
\begin{definition} \textup{(\textbf{Exclusivity})}
Exclusivity between two vectors $\bs{u}\in\mb{R}^m$ and $\bs{v}\in\mb{R}^m$ is defined as
$\mc{X}(\bs{u},\bs{v})\defeq\|\bs{u}\odot\bs{v}\|_0=\sum_{i=1}^m \bs{u}(i)\cdot\bs{v}(i)\neq 0$,
where $\odot$ designates the Hadamard product, and $\|\cdot\|_0$ is the $\ell^0$ norm.
\end{definition}
\noindent From the definition, we can observe that the exclusivity encourages two vectors to be as orthogonal as possible. Due to the non-convexity and discontinuity of $\ell^0$ norm, we have the following relaxed exclusivity. %, and thus can perform as a (powerful) diversity regularizer

\begin{definition} \textup{(\textbf{Relaxed Exclusivity})}
The definition of relaxed exclusivity between $\bs{u}\in\mb{R}^m$ and $\bs{v}\in\mb{R}^m$ is given as
$\mc{X}_r(\bs{u},\bs{v})\defeq\|\bs{u}\odot\bs{v}\|_1=\sum_{i=1}^m |\bs{u}(i)|\cdot|\bs{v}(i)|$,
where $|u|$ is the absolute value of $u$. The relaxation is similar with that of the $\ell^1$ norm to the $\ell^0$ norm.
\label{def:rlxexc}
\end{definition}
\noindent It can be easily verified that $\|\bs{u}\|_0=\mc{X}(\bs{u},\bs{1})$, $\|\bs{u}\|_1=\mc{X}_r(\bs{u},\bs{1})$ and $\|\bs{u}\|_2^2=\mc{X}_r(\bs{u},\bs{u})$, where $\bs{1}\in\mb{R}^m$ is the vector with all of its $m$ entries being $1$. 

Instead of directly using $\Psi(\bs{W}) \defeq\sum_{1\leq \tilde{c}\neq c\leq C}\mc{X}_r(\bs{w}_c,\bs{w}_{\tilde{c}})$, we employ the following:
\begin{equation}
\begin{aligned}
 \Psi(\bs{W}) \defeq\frac{1}{2}\|\bs{W}\|_F^2 +\sum_{1\leq \tilde{c}\neq c\leq C}\mc{X}_r(\bs{w}_c,\bs{w}_{\tilde{c}})=\frac{1}{2}\sum_{i=1}^M\bigg(\sum_{c=1}^C|\bs{w}_c(i)|\bigg)^2=\frac{1}{2}\|\bs{W}^T\|_{1,2}^2.
 \label{eq:reg}
 \end{aligned}
\end{equation}
The main reasons of bringing $\frac{1}{2}\|\bs{W}\|_F^2$ into the regularizer are: 1) it essentially enhances the stability of solution, 2) it tends to mitigate the scale issue by penalizing large columns, and 3) as the relaxed exclusivity itself is non-convex, the introduction guarantees the convexity of the regularizer. 
Finally, the proposed Exclusivity Regularized Machine (ERM) can be written in the following shape:
\begin{equation}
\begin{aligned}
\min_{\{\bs{w}_c,b_c\}} \frac{1}{2}\|\bs{W}^T\|_{1,2}^2
+\lambda\sum_{c=1}^C\sum_{i=1}^N \big(1-(\phi(\bs{x}_i)^T\bs{w}_c+b_c)y_i\big)_{+}^p.
 \end{aligned}
 \label{eq:erm}
\end{equation}
In next sub-section, we will customize an efficient and effective Augmented Lagrangian Multiplier algorithm (ALM) to seek the solution to \eqref{eq:erm}, which has rigorous convergence and global optimality guarantee as well as (quasi) linear complexity (discussed in Sec. \ref{sec:ta}). 

\textbf{Remarks} As expressed in Eq. \ref{eq:reg}, we have motivated the $\ell_{1,2}$ regularizer from a novel perspective. It has been verified that, as one of mixed norms, the $\ell_{1,2}$ is in nature able to capture some structured sparsity \cite{Mix}. In general, the regression models using such mixed norms can be solved by a modified FOCUSS algorithm \cite{Mix}. Zhou \textit{et al.} \cite{ExLasso} introduced the $\ell_{1,2}$ regularizer into a specific task, \textit{i.e.} multi-task feature selection, and used the subgradient method to seek the solution of the associated optimization problem. The responsibility of the $\ell_{1,2}$ regularizer is to enforce the negative correlation among categories \cite{ExLasso}. Recently, Kong \textit{et al.} \cite{Exc} utilized $\ell_{1,2}$ norm to bring out sparsity at intra-group level in feature selection, and proposed an effective iteratively re-weighted algorithm to solve the corresponding optimization problem. In this work, besides the view of motivating the $\ell_{1,2}$ regularizer, its role in our target problem, say constructing an ensemble of SVMs, is also different with the previous work \cite{Mix,ExLasso,Exc}. The functionalities of \cite{ExLasso} and \cite{Exc} are the intra-exclusivity of multiple hypotheses (tasks) and the inter-exclusivity of a single hypothesis respectively, while our principle is the diversity of multiple components of a single ensemble hypothesis. 

\subsection{Optimization}
With the trick that $1-(\phi(\bs{x}_i)^T\bs{w}_c+b_c)y_i=y_iy_i-(\phi(\bs{x}_i)^T\bs{w}_c+b_c)y_i=y_i(y_i-(\phi(\bs{x}_i)^T\bs{w}_c+b_c))$, we introduce auxiliary variables $e_i^c\defeq y_i-(\phi(\bs{x}_i)^T\bs{w}_c+b_c)$. In the sequel, the minimization of \eqref{eq:erm} can be converted into:
\begin{equation}
\begin{aligned}
&\argmin_{\{\bs{W},\bs{b}\}} \frac{1}{2}\|\bs{W}^T\|_{1,2}^2 
+\lambda \big(\bs{Y}\odot \bs{E})_{+}^p \quad
\subto\quad\bs{P} = \bs{W}; ~~ \bs{E} = \bs{Y}-(\bs{X}^T\bs{P}+\bs{1}\bs{b}^T),
 \end{aligned}
 \label{eq:erma}
\end{equation}
where $\bs{X}\in\mb{R}^{M\times N}\defeq[\phi(\bs{x}_1),\phi(\bs{x}_2),...,\phi(\bs{x}_N)]$, $\bs{e}^c\in\mb{R}^N\defeq[e_1^c,e_2^c,...,e_N^c]^T$, $\bs{E}\in\mb{R}^{N\times C}\defeq[\bs{e}^1,\bs{e}^2,...,\bs{e}^C]$ and $\bs{y}\in\mb{R}^N\defeq[y_1,y_2,...,y_N]^T$. And each column of $\bs{Y}\in\mb{R}^{N\times C}$ is $\bs{y}$. Please notice that, the constraint $\bs{P} = \bs{W}$ is added to make the objective separable and thus solvable by the Augmented Lagrangian Multiplier framework. It is worth mentioning that, thanks to the convexity of each term involved in the objective and the linearity of the constraints, the target problem is convex. The Lagrangian function of \eqref{eq:erma} can be written in the following form:
\begin{equation}
\begin{aligned}
\mc{L}(\bs{W},\bs{b},\bs{E},\bs{P})\defeq\frac{1}{2}\|\bs{W}^T\|_{1,2}^2 
+\lambda \big(\bs{Y}\odot \bs{E})_{+}^p+
\Phi(\bs{Q},\bs{P}-\bs{W})+\Phi(\bs{Z},\bs{E} -\bs{Y}+\bs{X}^T\bs{P}+\bs{1}\bs{b}^T),
 \end{aligned}
 \label{eq:erml}
\end{equation}
with the definition $\Phi(\bs{U},\bs{V})\defeq\frac{\mu}{2}\|\bs{V}\|_F^2+\langle\bs{U},\bs{V}\rangle$, where $\langle\cdot,\cdot\rangle$ represents matrix inner product and $\mu$ is a positive penalty scalar. In addition, $\bs{Q}\in\mb{R}^{M\times C}$ and $\bs{Z}\in\mb{R}^{N\times C}$ are Lagrangian multipliers. The proposed solver iteratively updates one variable at a time by fixing the others. Below are the solutions to sub-problems.

\noindent\textbf{$\bs{W}$ sub-problem} With the variables unrelated to $\bs{W}$ fixed, we have the sub-problem of $\bs{W}$:
\begin{equation}
\begin{aligned}
\bs{W}^{(t+1)} = \argmin_{\bs{W}} \frac{1}{2}\|\bs{W}^T\|_{1,2}^2 
+\Phi(\bs{Q}^{(t)},\bs{P}^{(t)}-\bs{W}).
\end{aligned}
\label{eq:W}
\end{equation}
As observed from the problem \eqref{eq:W}, it can be split into a set of smaller problems. For each row $\bs{W}_{\cdot j}$, instead of directly optimizing \eqref{eq:W}, we resolve the following equivalent objective:
\begin{equation}
\bs{W}^{(t+1)}_{\cdot j} = \argmin_{\bs{W}_{\cdot j}} \frac{1}{2} \bs{W}_{\cdot j}\bs{G}\bs{W}_{\cdot j}^T+\Phi(\bs{Q}^{(t)}_{\cdot j},\bs{P}^{(t)}_{\cdot j}-\bs{W}_{\cdot j}),
\label{eq:w}
\end{equation}
where $\bs{G}$ is formed by:
 \begin{equation}
 \bs{G}\defeq \diag\bigg(\bigg[\frac{\|\bs{W}_{\cdot j}\|_1}{|\bs{W}_{\cdot j}(1)|+\epsilon},\frac{\|\bs{W}_{\cdot j}\|_1}{|\bs{W}_{\cdot j}(2)|+\epsilon},...,\frac{\|\bs{W}_{\cdot j}\|_1}{|\bs{W}_{\cdot j}(C)|+\epsilon}\bigg]\bigg),
 \label{eq:G}
 \end{equation}
 where $\epsilon\rightarrow 0^+$ (a small constant) is introduced to avoid zero denominators.\footnote{The derived algorithm can be proved to minimize $\|\bs{W}^T+\epsilon\|_{1,2}^2$. Certainly, when $\epsilon\rightarrow 0^+$, $\|\bs{W}^T+\epsilon\|_{1,2}^2$ infinitely approaches to $\|\bs{W}^T\|_{1,2}^2$.}
 Since both $\bs{G}$ and $\bs{W}_{\cdot j}$ depend on $\bs{W}_{\cdot j}$, to find out the solution to \eqref{eq:w}, we employ an efficient re-weighted algorithm to iteratively update $\bs{G}$ and $\bs{W}_{\cdot j}$.
 As for $\bs{W}_{\cdot j}$, with $\bs{G}$ fixed, equating the partial derivative of \eqref{eq:w} with respect to $\bs{W}_{\cdot j}$ to zero yields:  
 \begin{equation}
\bs{W}_{\cdot j}^{(k+1)} = (\mu^{(t)}\bs{P}_{\cdot j}^{(t)}+\bs{Q}^{(t)}_{\cdot j})(\bs{G}^{(k)}+\mu^{(t)}\bs{I})^{-1}. 
 \label{eq:wk}
 \end{equation}
 Then $\bs{G}^{(k+1)}$ is updated using $\bs{W}_{\cdot j}^{(k+1)}$ as in \eqref{eq:G}. The procedure summarized in Algorithm \ref{alg:W} terminates when converged.
 
 \noindent\textbf{$\bs{b}$ sub-problem} Dropping the terms independent on $\bs{b}$ leads to a least squares regression problem:
 \begin{equation}
\begin{aligned}
\bs{b}^{(t+1)} = \argmin_{\bs{b}} \Phi(\bs{Z}^{(t)},\bs{E}^{(t)} -\bs{Y}+\bs{X}^T\bs{P}^{(t)}+\bs{1}\bs{b}^T)= \big(\bs{Y}-\bs{E}^{(t)}-\bs{X}^T\bs{P}^{(t)}-\frac{\bs{Z}^{(t)}}{\mu^{(t)}}\big)^T\big(\frac{1}{N}\bs{1}\big).
\end{aligned}
\label{eq:bs}
\end{equation}
%The closed-form solution to $\bs{b}$ can be easily obtained via:
%\begin{equation}
%\begin{aligned}
%\bs{b}^{(t+1)} = \big(\bs{Y}-\bs{E}^{(t)}-\bs{X}^T\bs{P}^{(t)}-\frac{\bs{Z}^{(t)}}{\mu^{(t)}}\big)^T\big(\frac{1}{N}\bs{1}\big).
%\end{aligned}
%\label{eq:bs}
%\end{equation}

%& \argmin_{\bs{E}}\lambda \big(\bs{Y}\circ \bs{E})_{+}^p+\Phi(\bs{Z}^{(t)},\bs{E} -\bs{Y}+\bs{X}^T\bs{P}^{(t)}+\bs{1}\bs{b}^{(t+1)T})\\=&
\noindent\textbf{$\bs{E}$ sub-problem} Similarly, picking out the terms related to $\bs{E}$ gives the following problem:
\begin{equation}
\begin{aligned}
\bs{E}^{(t+1)} =\argmin_{\bs{E}}\frac{\lambda}{\mu^{(t)}} \big(\bs{Y}\circ \bs{E})_{+}^p+\frac{1}{2}\|\bs{E}-\bs{S}^{(t)}\|_F^2,
\end{aligned}
\label{eq:E}
\end{equation}
where $\bs{S}^{(t)}\defeq\bs{Y}-\bs{X}^T\bs{P}^{(t)}-\bs{1b}^{(t+1)T}-\frac{\bs{Z}^{(t)}}{\mu^{(t)}}$.
It can be seen that the above is a single-variable $2$-piece piecewise function. Thus, to seek the minimum of each element in $\bs{E}$, we just need to pick the smaller between the minima when $y_ie_i^c\geq 0$ and $y_ie_i^c< 0$. Moreover, we can provide the explicit solution when $p\defeq 1$ or $2$ (for arbitrary $p$ we will discuss it in Sec. \ref{sec:c&d}). When $p\defeq 1$:
\begin{equation}
\bs{E}^{(t+1)} = \bs{\Omega}\circ\mc{S}_{\frac{\lambda}{\mu^{(t)}}}[\bs{S}^{(t)}]+\bs{\bar{\Omega}}\circ\bs{S}^{(t)}.
\label{eq:e1}
\end{equation}
For $p\defeq 2$:
\begin{equation}
\bs{E}^{(t+1)} = \bs{\Omega}\circ{\bs{S}^{(t)}}/({1+\frac{2\lambda}{\mu^{(t)}}})+\bs{\bar{\Omega}}\circ\bs{S}^{(t)},
\label{eq:e2}
\end{equation}
where $\bs{\Omega}\in\mb{R}^{N\times C}\defeq(\bs{Y}\circ\bs{S}^{(t)}>0)$ is an indicator matrix, and $\bs{\bar{\Omega}}$ is the complementary support of $\bs{{\Omega}}$.  The definition of shrinkage operator on scalars is $\mathcal{S}_{\epsilon>0}[u]\defeq\sgn(u)\max(|u|-\epsilon,0)$. The extension of the shrinkage operator to vectors and matrices is simply applied element-wise.

\begin{algorithm}[t]
\SetAlgoLined \caption{$\bs{W}$ Solver}
\KwIn{$\bs{W}^{(t)}$, $\bs{P}^{(t)}$, $\bs{Q}^{(t)}$, $\mu^{(t)}$.  \textbf{Initial.:} $k\leftarrow 0$; $\bs{H}^{(k)}\leftarrow\bs{W}^{(t)}$;}  
\For{$j=0:M$}{
\While{not converged}{
Update $\bs{G}^{(k+1)}$ via Eq. \eqref{eq:G}; 
Update $\bs{H}^{(k+1)}_{\cdot j}$ via Eq. \eqref{eq:wk};
$k\leftarrow k+1$;\\
}
}
\KwOut{$\bs{W}^{(t+1)}\leftarrow\bs{H}^{(k)}$}
\label{alg:W}
\end{algorithm}

\begin{algorithm}[t]
\SetAlgoLined \caption{Exclusivity Regularized Machine}
\KwIn{Training set $\{(\bs{x}_i, y_i)\}_{i=1}^N$, positive integer $C$ and positive real value $\lambda$.}
\textbf{Initial.:} $t\leftarrow 0$; $\bs{W}^{(t)}\in\mathbb{R}^{M\times C}\leftarrow \bs{1}$; $\bs{b}^{(t)}\in\mb{R}^{C}\leftarrow\bs{0}$; $\bs{P}^{(t)}\in\mathbb{R}^{M\times C}\leftarrow \bs{0}$;$\bs{Q}^{(t)}\in\mathbb{R}^{M\times C}\leftarrow \bs{1}$; $\bs{Z}^{(t)}\in\mathbb{R}^{N\times C}\leftarrow \bs{0}$; $\mu^{(t)}\leftarrow 1$; $\rho\leftarrow 1.1$; \\
\While{not converged}{
Update $\bs{W}^{(t+1)}$ via Alg. \ref{alg:W}; Update $\bs{b}^{(t+1)}$ via Eq. \eqref{eq:bs}; 
Update $\bs{E}^{(t+1)}$ via Eq. \eqref{eq:e1}  or \eqref{eq:e2}; \\Update $\bs{P}^{(t+1)}$ via Eq. \eqref{eq:Ps}; 
Update Multipliers and $\mu^{(t+1)}$ via Eq. \eqref{eq:mul}; $t\leftarrow t+1$;\\
}
\KwOut{Final Ensemble $\frac{1}{C}(\sum_{i=1}^C\bs{w}_c^{(t)},\sum_{i=1}^C{b}_c^{(t)})$}
\label{alg:ERM}
\end{algorithm}

\noindent\textbf{$\bs{P}$ sub-problem} There are two terms involve $\bs{P}$. The associated optimization problem reads:
\begin{equation}
\begin{aligned}
\bs{P}^{(t+1)} =& \argmin_{\bs{P}} \Phi(\bs{Q}^{(t)},\bs{P}-\bs{W}^{(t+1)})+\Phi(\bs{Z}^{(t)},\bs{E}^{(t+1)} -\bs{Y}+\bs{X}^T\bs{P}+\bs{1}\bs{b}^{(t+1)T}).
\end{aligned}
\label{eq:P}
\end{equation}
This sub-problem only contains quadratic terms, so it is easy to compute the solution in closed-form:
\begin{equation}
\begin{aligned}
\bs{P}^{(t+1)} =
\bs{K}^{-1}\big(\bs{W}^{(t+1)}-\frac{\bs{Q}^{(t)}}{\mu^{(t)}}+\bs{X}(\bs{M}-\bs{E}^{(t+1)})\big),
\end{aligned}
\label{eq:Ps}
\end{equation}
where we denote $\bs{K}^{-1}\defeq(\bs{I}+\bs{XX}^T)^{-1}$ and $\bs{M}\defeq\bs{Y}-\bs{1b}^{(t+1)T}-\frac{\bs{Z}^{(t)}}{\mu^{(t)}}$.

\noindent\textbf{Multipliers and $\mu$} Besides, there are two multipliers and $\mu$ to update, which are simply given by:
\begin{equation}
\begin{aligned}
\bs{Z}^{(t+1)}&=\bs{Z}^{(t)}+\mu^{(t)}(\bs{E}^{(t+1)} -\bs{Y}+\bs{X}^T\bs{P}^{(t+1)}+\bs{1}\bs{b}^{(t+1)T});\\
\bs{Q}^{(t+1)} &= \bs{Q}^{(t)}+\mu^{(t)}(\bs{P}^{(t+1)}-\bs{W}^{(t+1)}); \mu^{(t+1)}=\rho\mu^{(t)}, \rho>1.
\end{aligned}
\label{eq:mul}
\end{equation}

For clarity, the procedure of solving \eqref{eq:svmh} is outlined in Algorithm \ref{alg:ERM}. The algorithm should not be terminated until the change of objective value is smaller than a pre-defined threshold (in the experiments, we use $0.05$). Please see Algorithm \ref{alg:ERM} for other details that we can not cover in the text.

\section{Theoretical Analysis}
\label{sec:ta}
First, we come to the loss term of ERM \eqref{eq:erm}, which accesses the total penalty of base learners as:
\begin{equation}
\sum_{c=1}^C\sum_{i=1}^N \big(1-(\phi(\bs{x}_i)^T\bs{w}_c+b_c)y_i\big)_{+}^p,
\end{equation}
where $p\geq 1$. Alternatively, the loss of the ensemble $\{\bs{w}_e,b_e\}\defeq\{\frac{1}{C}\sum_{c=1}^C\bs{w}_c, \frac{1}{C}\sum_{c=1}^Cb_c\}$ is as:
\begin{equation}
\sum_{i=1}^N \big(1-(\phi(\bs{x}_i)^T\bs{w}_e+b_e)y_i\big)_{+}^p.
\label{eq:ensL}
\end{equation}
Based on the above, we have the relationship between the two losses as described in Proposition \ref{pro:upb}.
\begin{proposition} 
Let $\{\bs{w}_1, b_1\}$,..., $\{\bs{w}_C, b_C\}$ be the component learners obtained by ERM (Alg. \ref{alg:ERM}), and $\{\bs{w}_e,b_e\}\defeq\{\frac{1}{C}\sum_{c=1}^C\bs{w}_c, \frac{1}{C}\sum_{c=1}^Cb_c\}$ the ensemble, the loss of $\{\bs{w}_e,b_e\}$ is bounded by the average loss of the base learners. 
\label{pro:upb}
\end{proposition}
\vspace{-10pt}
\begin{proof}
Please note that, for each training instance, substituting $\{\bs{w}_e,b_e\}$ with $\{\frac{1}{C}\sum_{c=1}^C\bs{w}_c, \frac{1}{C}\sum_{c=1}^Cb_c\}$ into \eqref{eq:ensL} yields $\sum_{i=1}^N\big(\frac{1}{C}\sum_{c=1}^C\big(1-(\phi(\bs{x}_i)^T\bs{w}_c+b_c)y_i\big)\big)_{+}^p$. Due to the convexity of the hinge loss together with $p\geq 1$, the relationship $\sum_{i=1}^N \big(1-(\phi(\bs{x}_i)^T\bs{w}_e+b_e)y_i\big)_{+}^p
\leq \frac{1}{C}\sum_{c=1}^C\sum_{i=1}^N \big(1-(\phi(\bs{x}_i)^T\bs{w}_c+b_c)y_i\big)_{+}^p$ holds by applying the Jensen's inequality.
\end{proof}
\noindent The proposition indicates that as we optimize ERM \eqref{eq:erm}, an upper bound of the loss of the ensemble is also minimized. Thus, incorporating with our proposed regularizer, ERM is able to achieve the goal of simultaneously optimizing the training error of ensemble and the diversity of components. 

Next, we shall consider the convergence and optimality of the designed algorithms. Before discussing Alg. \ref{alg:ERM}, we have to confirm the property of Alg. \ref{alg:W}, which is established by Theorem \ref{the:W}. 

\begin{theorem}
The updating rules \eqref{eq:G} and \eqref{eq:wk} for solving the problem \eqref{eq:w}, \textit{i.e.} Algorithm \ref{alg:W}, converges and the obtained optimal solution is exactly the global optimal solution of the problem \eqref{eq:W}.
\label{the:W}
\end{theorem}
\vspace{-10pt}
\begin{proof}
Algorithm \ref{alg:W} is actually a special case of the algorithm proposed in \cite{Exc}. Due to the limited space, we refer readers to \cite{Exc} for the detailed proof. 
\end{proof}

\noindent With Theorem \ref{the:W}, it is ready to analyze Algorithm \ref{alg:ERM}. To this end, the following lemmas are required.
\begin{lemma} \cite{MRM,ALMM} 
Let $\mathcal{H}$ be a real Hilbert space endowed with an inner product $\langle\cdot,\cdot\rangle$ and a corresponding norm $\|\cdot\|$, and any $\bs{y}\in\partial\|\bs{x}\|$, where $\partial\|\cdot\|$ denotes the subgradient. Then $\|\bs{y}\|^*=1$ if $\bs{x}\neq 0$, and $\|\bs{y}\|^*\leq 1$ if $\bs{x}= 0$, where $\|\cdot\|^*$ is the dual norm of the norm $\|\cdot\|$.
\label{lemma:dual}
\end{lemma}

\begin{lemma} Both the sequences $\{\bs{Z}^{(t)}\}$ and $\{\bs{Q}^{(t)}\}$ generated by Algorithm \ref{alg:ERM} are bounded.
\label{lemma:muliplers}
\end{lemma}
\vspace{-10pt}
\begin{proof}
According to the optimality conditions for \eqref{eq:erma} with respect to $\bs{E}$ and $\bs{W}$, and the updating rules of multipliers \eqref{eq:mul}, we know
$\bs{Z}^{(t+1)}\in\partial\|\lambda(\bs{Y}\circ\bs{E}^{(t+1)})_+\|_{p}^p;-\bs{Q}^{(t+1)}\in\partial\|\frac{1}{2}\bs{W}^T\|^2_{1,2}$.
Using Lemma \ref{lemma:dual} reaches that the sequences $\{\bs{Z}^{(t)}\}$ and $\{\bs{Q}^{(t)}\}$ are both bounded because the dual norms of $\|\cdot\|_p$ and $\|\cdot\|_{1,2}$ are $\|\cdot\|_{\frac{p}{p-1}}$ and $\|\cdot\|_{\infty,2}$, respectively.
%\begin{equation}
%\|\bs{Z}^{(t+1)}\|_{\frac{p}{p-1}}\leq\lambda.
%\end{equation}
%Thus, the sequence $\{\bs{Z}^{(t)}\}$ is bounded. The boundedness of the sequence $\{\bs{Q}^{(t)}\}$ can be proved in a similar way by taking derivative of \eqref{eq:erma} with respect to $\bs{W}$.
\end{proof}

Now, we have come to the convergence and optimality of our proposed Algorithm \ref{alg:ERM}.
\begin{theorem}
The solution consisting of the limit of the sequences $\{\bs{W}^{(t)}\}$, $\{\bs{b}^{(t)}\}$ and $\{\bs{E}^{(t)}\}$ generated by Algorithm \ref{alg:ERM}, \textit{i.e.} $(\bs{W}^{(\infty)}, \bs{b}^{(\infty)},\bs{E}^{(\infty)})$, is global optimal to ERM \eqref{eq:erm}, and the convergence rate is at least $o(\frac{1}{\mu^{(t)}})$.
\label{the:ERM}
\end{theorem}
\vspace{-10pt}
\begin{proof}
As the vital natural property of an ALM algorithm, the following holds:
 \begin{equation}
 \begin{aligned}
 &\mc{L}_{\mu^{(t)}}(\bs{W}^{(t+1)},\bs{b}^{(t+1)},\bs{E}^{(t+1)},\bs{P}^{(t+1)},\bs{Q}^{(t)},\bs{Z}^{(t)})
 \leq  \mc{L}_{\mu^{(t)}}(\bs{W}^{(t)},\bs{b}^{(t)},\bs{E}^{(t)},\bs{P}^{(t)},\bs{Q}^{(t)},\bs{Z}^{(t)})=\\
  & \mc{L}_{\mu^{(t-1)}}(\bs{W}^{(t)},\bs{b}^{(t)},\bs{E}^{(t)},\bs{P}^{(t)},\bs{Q}^{(t-1)},\bs{Z}^{(t-1)})+\frac{1+\rho}{2\mu^{(t-1)}}(\|\bs{Q}^{(t)}-\bs{Q}^{(t-1)}\|_F^2+\|\bs{Z}^{(t)}-\bs{Z}^{(t-1)}\|_F^2).
 \end{aligned}
 \label{eq:rel}
 \end{equation}
 Due to $\sum_{t=1}^{\infty} \frac{1+\rho}{2\mu^{(t-1)}}=\frac{\rho(1+\rho)}{2\mu^{(0)}(\rho-1)}<\infty$ and the boundedness of $\{\bs{Q}^{(t)}\}$ and $\{\bs{Z}^{(t)}\}$,
 we can conclude that $\mc{L}_{\mu^{(t-1)}}(\bs{W}^{(t)},\bs{b}^{(t)},\bs{E}^{(t)},\bs{P}^{(t)},\bs{Q}^{(t-1)},\bs{Z}^{(t-1)})$ is upper bounded. As a consequence,  
 \begin{equation}
 \begin{aligned}
\frac{1}{2}\|\bs{W}^{(t)}\|_{1,2}^2+\lambda\|(\bs{Y}\circ\bs{E}^{(t)})_+\|_p^p=&\mc{L}_{\mu^{(t-1)}}(\bs{W}^{(t)},\bs{b}^{(t)},\bs{E}^{(t)},\bs{P}^{(t)},\bs{Q}^{(t-1)},\bs{Z}^{(t-1)})-\\
&\frac{\|\bs{Q}^{(t)}\|_F^2-\|\bs{Q}^{(t-1)}\|_F^2}{2\mu^{(t-1)}}-\frac{\|\bs{Z}^{(t)}\|_F^2-\|\bs{Z}^{(t-1)}\|_F^2}{2\mu^{(t-1)}}
 \end{aligned}
 \label{eq:opt}
  \end{equation}
is upper bounded, that is to say $\{\bs{W}^{(t)}\}$ and $\{\bs{E}^{(t)}\}$ are bounded. According to the updating rules of multipliers, the constraints are satisfied when $t\rightarrow\infty$. In other words, due to the boundedness of the sequences $\{\bs{Q}^{(t)}\}$ and $\{\bs{Z}^{(t)}\}$, the right sides of $\bs{E}^{(t+1)} -\bs{Y}+\bs{X}^T\bs{P}^{(t+1)}+\bs{1}\bs{b}^{(t+1)T}=\frac{\bs{Z}^{(t+1)}-\bs{Z}^{(t)}}{\mu^{(t)}}$ and $\bs{P}^{(t+1)}-\bs{W}^{(t+1)} = \frac{\bs{Q}^{(t+1)}-\bs{Q}^{(t)}}{\mu^{(t)}}$ infinitely approximate to $0$. This proves the feasibility of the solution by Alg. \ref{alg:ERM} as well as the boundedness of $\{\bs{P}^{(t)}\}$ and $\{\bs{b}^{(t)}\}$.
  
Because of the above \eqref{eq:opt} and the boundedness of multipliers $\{\bs{Q}^{(t)}\}$ and $\{\bs{Z}^{(t)}\}$, we have:
\begin{equation}
\begin{aligned}
&\lim_{t\rightarrow\infty} \frac{1}{2}\|\bs{W}^{(t)}\|_{1,2}^2+\lambda\|(\bs{Y}\circ\bs{E}^{(t)})_+\|_p^p = \lim_{t\rightarrow\infty} \mc{L}_{\mu^{(t-1)}}(\bs{W}^{(t)},\bs{b}^{(t)},\bs{E}^{(t)},\bs{P}^{(t)},\bs{Q}^{(t-1)},\bs{Z}^{(t-1)}).
\end{aligned}
\end{equation}
Thanks to the feasibility of solution, the last two terms in \eqref{eq:erma} are neglectable. So Alg. \ref{alg:ERM} converges to a global optimal solution to \eqref{eq:erm}. The convergence rate is at least $o(\frac{1}{\mu^{(t)}})$ according to \eqref{eq:opt}. 
\end{proof}

In addition, we show the complexity of Alg. \ref{alg:ERM}. The operations including matrix addition and subtraction are relatively cheap, and thus can be ignored. Updating each row of $\bs{W}$ takes $\mc{O}(qC^2)$ and $\mc{O}(qC)$ for \eqref{eq:wk} and \eqref{eq:G} respectively, where $q$ is the (inner) iteration number of Alg. \ref{alg:W}. Please note that, due to the diagonality of $\bs{G}$, the inverse of $\bs{G}+\mu\bs{I}$ only needs $\mc{O}(C)$. Therefore, the cost of Alg. \ref{alg:W} is $\mc{O}(qC^2M)$. The $\bs{b}$ sub-problem requires $\mc{O}(CMN)$. The complexity of the $\bs{E}$ sub-problem is $\mc{O}(CMN)$, for both $p\defeq 1$ and $p\defeq 2$. Solving $\bs{P}$ spends $\mc{O}(CMN+CM^2)$. Besides, the update of the multipliers is at $\mc{O}(CMN)$ expense. In summary, Alg. \ref{alg:ERM} has the complexity of $\mc{O}(tCM(qC+N+M))$, where $t$ is the number of (outer) iterations required to converge.

\section{Experimental Verification}
We use $9$ popularly adopted benchmark datasets from various sources for performance evaluation: including \textit{sonar} $(N=208, M=60)$, \textit{german} $(1,000, 24)$, \textit{australian} $(690, 14)$, \textit{ijcnn1} $(49,990, 22)$, \textit{heart} $(270, 13)$, \textit{ionosphere} $(351, 34)$, \textit{diabetes} $(768, 8)$, \textit{liver} $(345, 6)$ and \textit{splice} $(1,000, 60)$.\footnote{All available at \text{www.csie.ntu.edu.tw/$\sim$cjlin/libsvmtools/datasets}} All experiments are conducted on a machine with $2.5$ GHz CPU and $64$G RAM.

\begin{figure} \centering
		\includegraphics[width=0.25\linewidth]{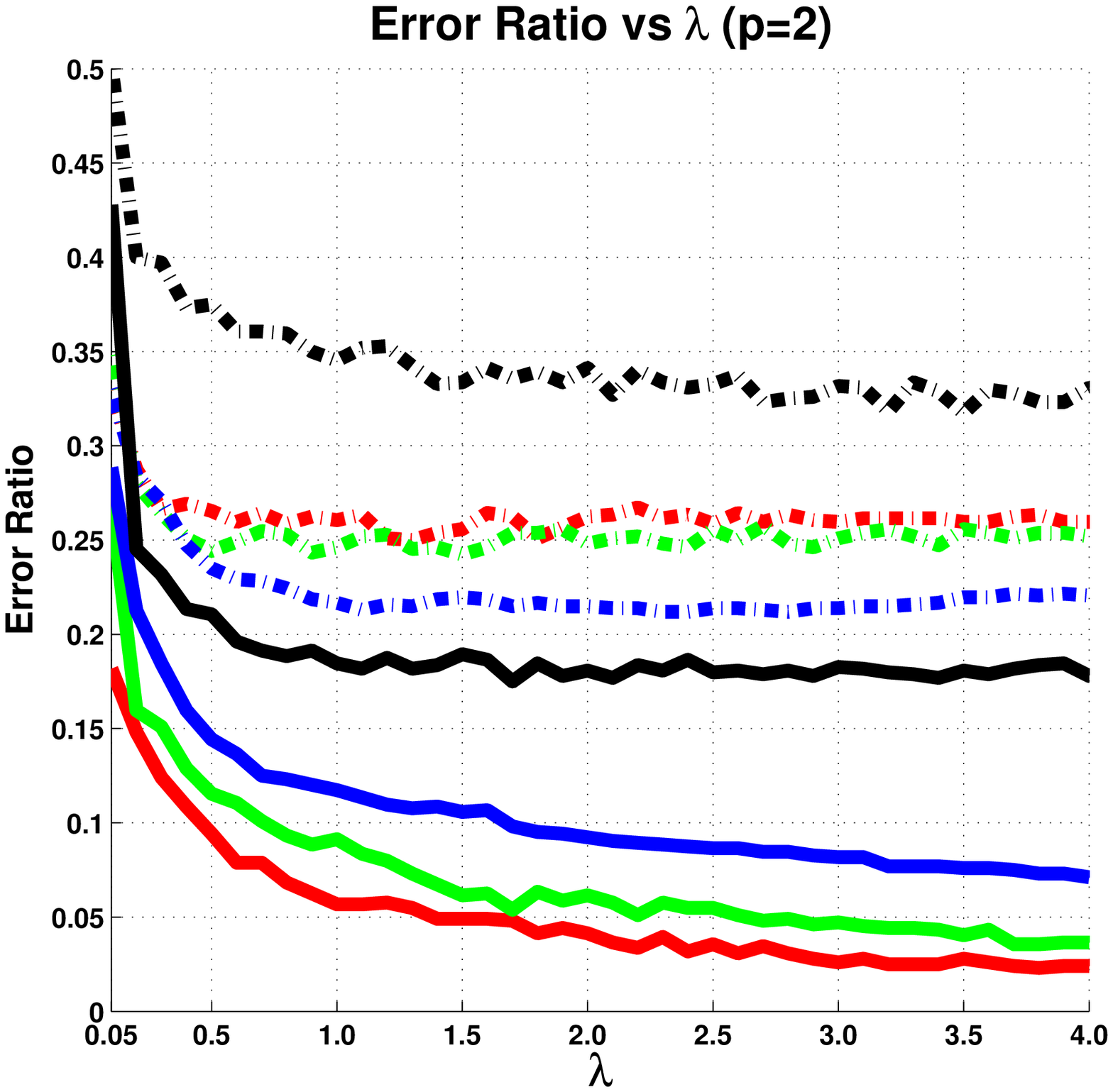}\hspace{-5pt}
		\includegraphics[width=0.25\linewidth]{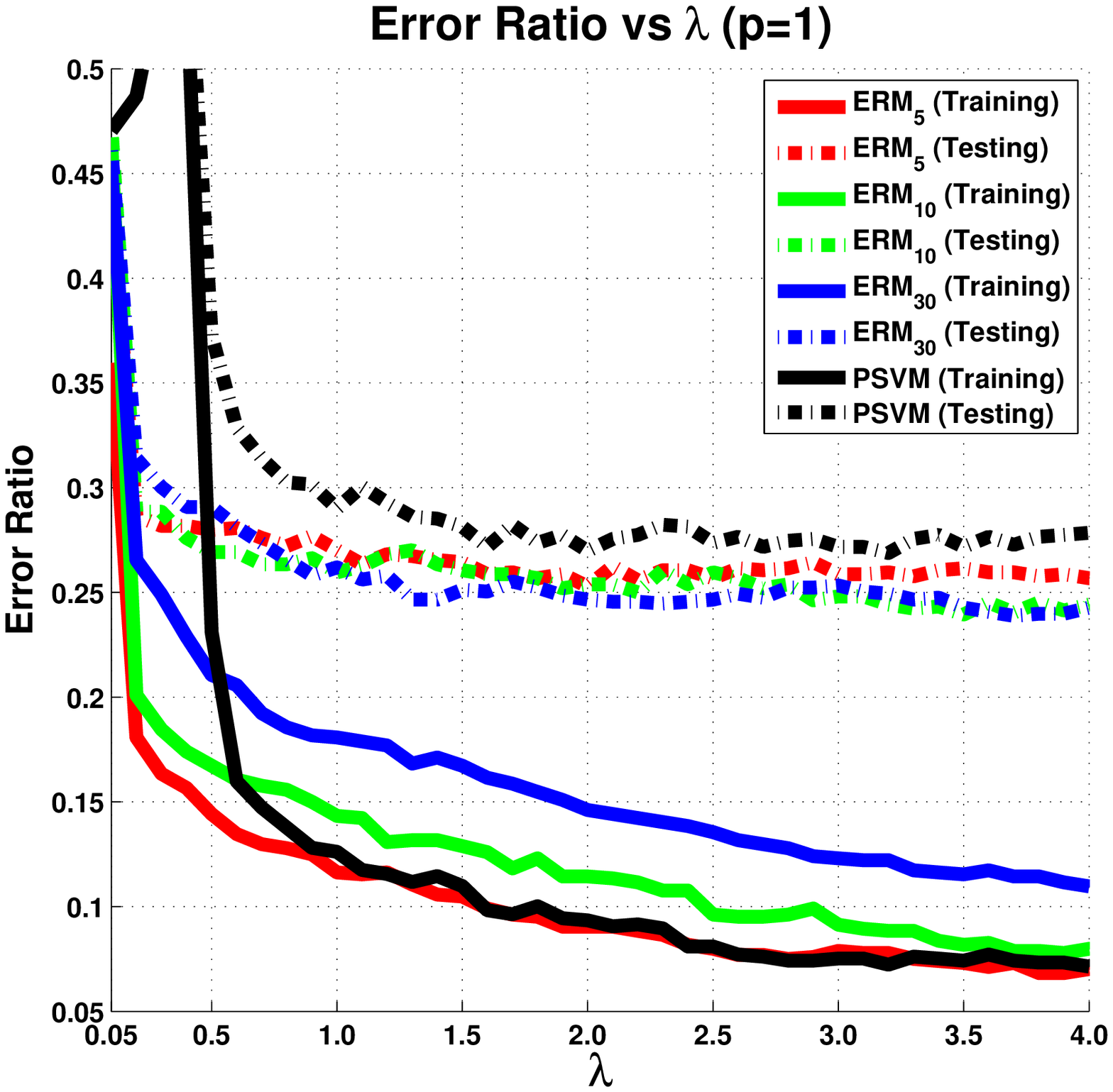}\hspace{-5pt}
		\includegraphics[width=0.25\linewidth]{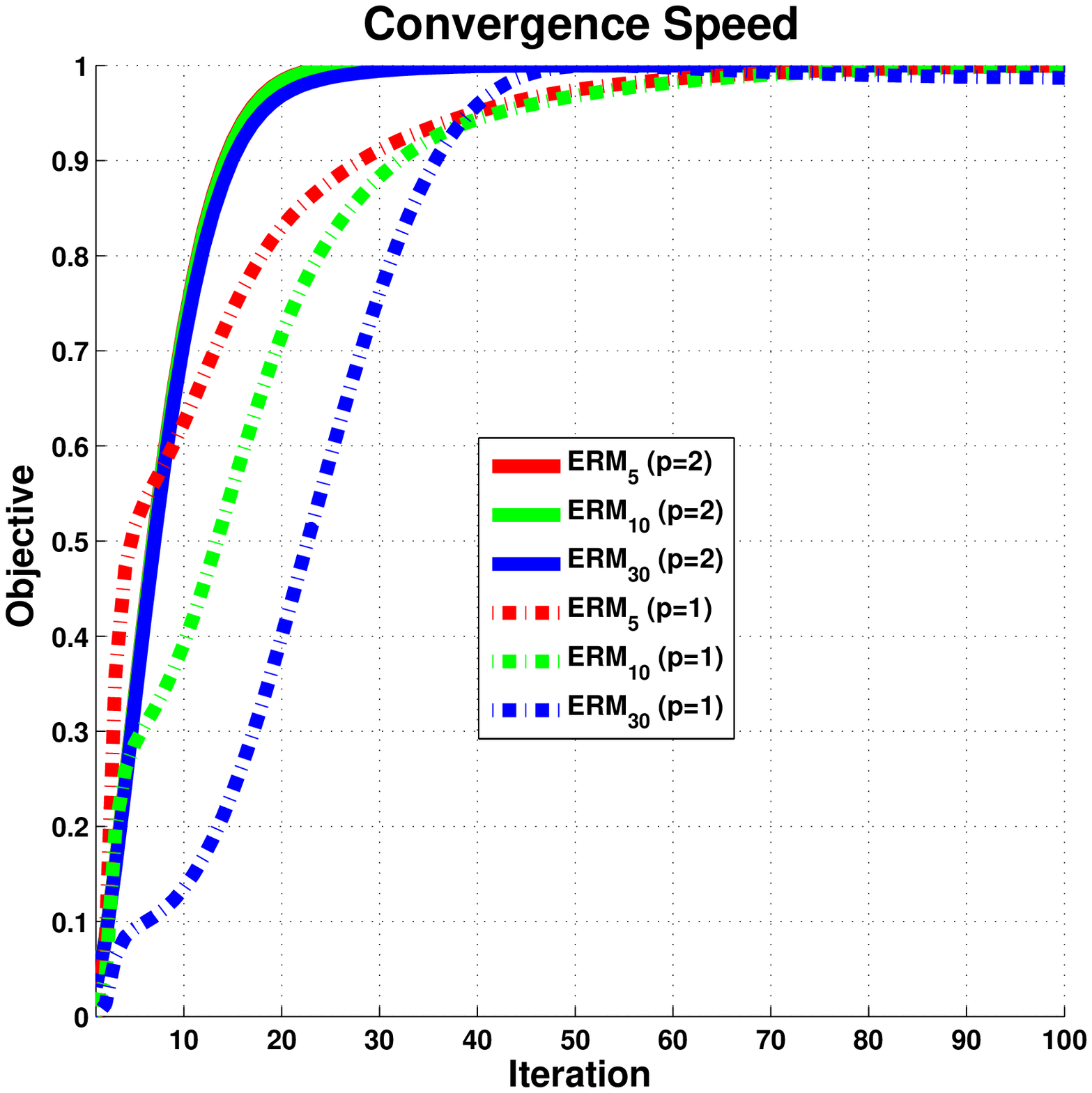}\hspace{-5pt}
		\includegraphics[width=0.25\linewidth]{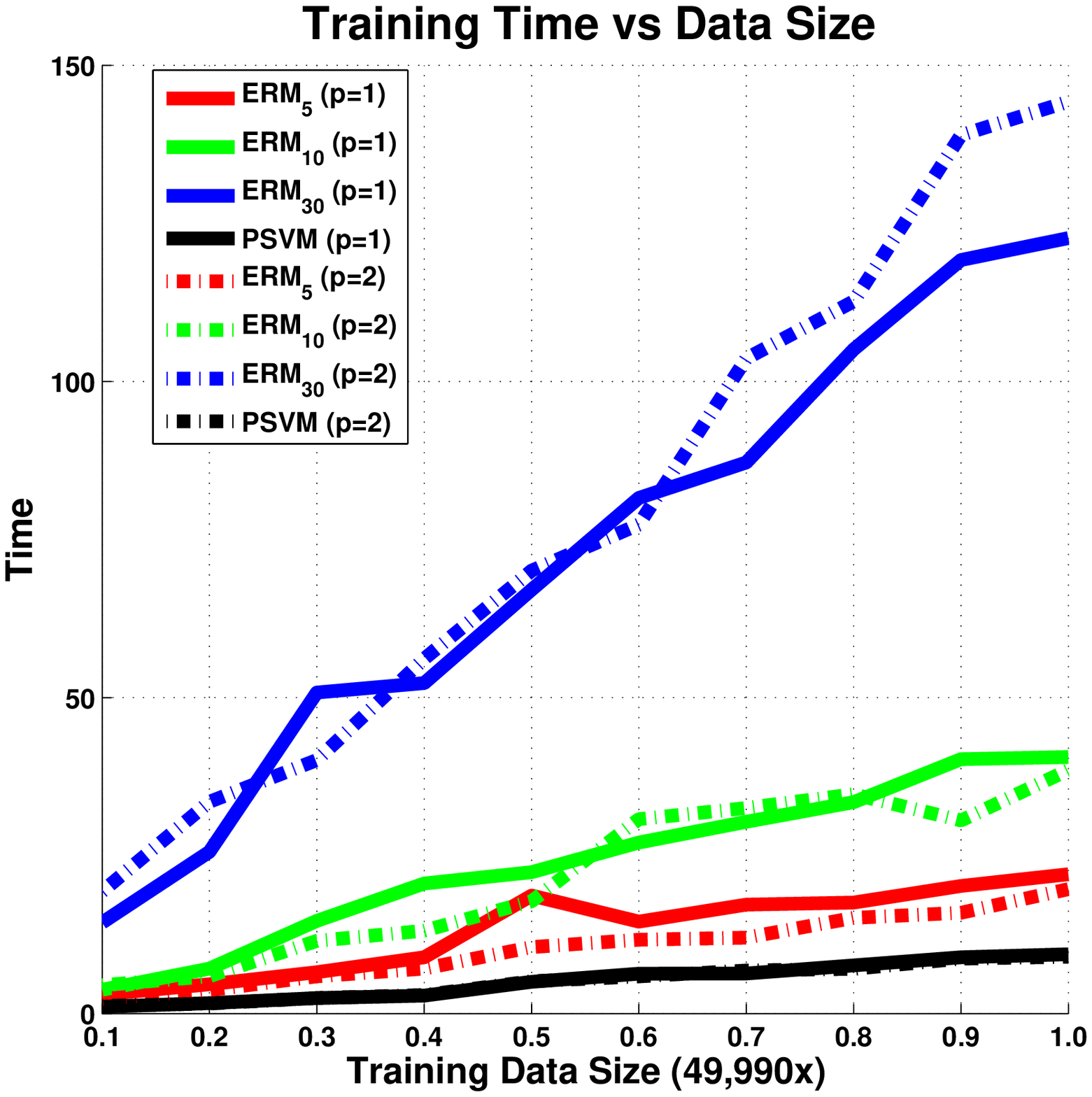}
		\vspace{-10pt}
	\caption{Parameter effect of $\lambda$, convergence speed and training time}
	\label{fig:pe}
	\vspace{-10pt}
\end{figure}

\textbf{Parameter Effect}
Here, we evaluate the training and testing errors of ERM$_C$ ($C\in\{5,10,30\}$ means the component number) against varying values of  $\lambda$ in the range $[0.05, 4]$. All the results shown in this experiment are averaged over $10$ independent trials, each of which randomly samples half data from the \textit{sonar} dataset for training and the other half for testing. The first picture in Fig. \ref{fig:pe} displays the training error and testing error plots of L$_2$ loss ERM with L$_2$ loss PSVM \cite{PSVM-ALM} (denoted as L$_2$-PSVM) as reference. From the curves, we can observe that, as $\lambda$ grows, the training errors drop, as well as composing less base learns leads to a smaller training error. This is because more and more effort is put on fitting data. As regards the testing error, the order is reversed, which corroborates the recognition that the predication gains from the diversity of classifiers, and reveals the effectiveness of our design in comparison with L$_2$-PSVM. Besides, the testing errors change very slightly in a relatively large range of $\lambda$, which implies the insensitivity of ERM to $\lambda$. The second picture corresponding to $p\defeq 1$ shows an additional evidence to $p\defeq 2$.  Although the performance gaps between the different cases shrink, the improvement of ERM is still noticeable. Based on this evaluation, we set $\lambda$ to $2$ for both L$_1$-ERM and L$_2$-ERM in the rest experiments.

\textbf{Convergence Speed \& Training Time }
\label{sec:c&t}
%\begin{figure} \centering
%		\includegraphics[width=0.48\linewidth]{./figure/convergence.eps}
%		\includegraphics[width=0.48\linewidth]{./figure/time.eps}
%	\caption{Convergence speed and training time}
%	\vspace{-4pt}
%	\label{fig:CT}
%\end{figure}
Although the convergence rate and complexity of the proposed algorithm have been theoretically provided, it would be more intuitive to see its empirical behavior. Thus, we here show how quick the algorithm converges, without loss of generality, on the \textit{ijcnn1} dataset. From the third picture in Fig. \ref{fig:pe}, we can observe that, when $p\defeq2$, all the three cases converge with about $30$ iterations. The cases correspond to $p\defeq1$ take more iterations than $p\defeq2$ (about $70$ iterations), but they are still very efficient. Please note that, for a better view of different settings, the objective plots are normalized into the range $[0, 1]$.
The most right picture in Fig. \ref{fig:pe} gives curves of how the CPU-time used for training increases with respect to the number of training samples. Since the training time is too short to be accurately counted, we carry out each test for $10$ independent trials, and report the total training time (in seconds). As can be seen, the training time for both $p\defeq 1$ and $2$ is quasi linear with respect to the size of training data. For all the three cases correspond to ERM$_5$, ERM$_{10}$ and ERM$_{30}$, the choice of $p$ barely brings differences in time. The gaps between the three cases dominantly come from the number of base learners. The primal SVM only needs to learn one classifier while ERM requires to train multiple bases.\footnote{In \cite{PSVM-ALM}, the authors have revealed via extensive experiments, that PSVM (SVM-ALM) is much more efficient than SVM$^{perf}$ \cite{SVM-perf}, Pegasos \cite{Pegasos}, BMRM \cite{BMRM}, and TRON \cite{TRON}, PCD \cite{PCD} and DCD \cite{DCD}. } 

\begin{table*}[t]
\caption{Testing errors (mean $\pm$ standard deviation, \%) on benchmark datasets. }	
\vspace{-10pt}
	\label{tab:acc}
	\begin{center}
	\resizebox{\textwidth}{!}{
		\begin{tabular}{c|| c c c c c c c c ||c}
		   \hline
		   \textbf{Method} (R) & \textbf{\textit{german}} & \textbf{\textit{diabetes}} & \textbf{\textit{australian}} & \textbf{\textit{sonar}} & \textbf{\textit{splice}}  & \textbf{\textit{liver}}  & \textbf{\textit{heart}} & \textbf{\textit{ionosphere}}  & \textbf{{A.R.}} \\
		   \hline
          { \textbf{L$_1$-ERM$_{10}$}}   & 26.08$\pm$1.21 (5) & 24.73$\pm$1.44 (4) & 14.59$\pm$0.84 (6) & 23.62$\pm$3.64 (4) & 26.53$\pm$1.66 (7) & 42.82$\pm$2.41 (6) & 17.17$\pm$1.25 (2)  & 13.68$\pm$2.60 (4)& 4.8       \\

            \hline
            
           { \textbf{ L$_2$-ERM$_{10}$}}   & 26.00$\pm$1.16 (3) & 24.34$\pm$0.92 (1)& 14.13$\pm$0.40 (3) & 23.79$\pm$4.36 (5) & 26.75$\pm$2.26 (8) & 36.00$\pm$4.05 (1)& 17.83$\pm$1.81 (4) & 13.03$\pm$2.20 (2)& 3.4    \\

		   \hline
            {\textbf{AdaBoost$_{10}$} }      & 30.51$\pm$1.38 (8) & 29.24$\pm$2.73 (7)& 48.39$\pm$1.91 (9) & 27.59$\pm$6.08 (10) & 14.86$\pm$3.03 (3) & 46.92$\pm$3.59 (10)& 47.83$\pm$3.25 (9)  & 30.05$\pm$3.13 (8) & 8.0        \\

           \hline
            {\textbf{ Bagging$_{10}$} }         & 31.59$\pm$2.59 (9) & 30.63$\pm$3.38 (9) & 46.43$\pm$2.58 (7)& 25.00$\pm$8.26 (6)& 19.26$\pm$1.78 (4) & 46.00$\pm$3.58 (8)& 47.58$\pm$1.94 (7) & 38.71$\pm$3.02 (10) & 7.5       \\

           \hline
            { \textbf{DRM$_{10}$ }  }            & 25.98$\pm$1.20 (2) & 24.47$\pm$0.89 (2) & 14.09$\pm$1.55 (2) & 27.07$\pm$3.05 (9)& 35.96$\pm$2.85 (9) & 36.77$\pm$3.79 (3) & 19.00$\pm$2.22 (5) & 19.50$\pm$3.67 (5) & 4.6      \\

            \hline
          {\textbf{ L$_1$-ERM$_{30}$} }    & 26.27$\pm$1.02 (6) & 33.50$\pm$1.92 (10) & 14.24$\pm$1.02 (4)& 23.62$\pm$2.82 (3)& 25.64$\pm$1.63 (5)& 42.77$\pm$2.48 (5) & 17.17$\pm$2.30 (3) & 13.30$\pm$2.17 (3) & 4.9         \\

		   \hline
         {\textbf{ L$_2$-ERM$_{30}$} }    & 25.75$\pm$1.10 (1) & 25.42$\pm$1.41 (5) & 14.02$\pm$0.81 (1) & 21.55$\pm$3.66 (1) & 26.07$\pm$1.80 (6) & 40.05$\pm$3.48 (4) & 17.08$\pm$1.26 (1) & 12.99$\pm$1.95 (1) & 2.5       \\

           \hline
           { \textbf{AdaBoost$_{30}$ } }    & 32.22$\pm$2.32 (10)& 29.85$\pm$2.77 (8) & 48.94$\pm$2.13 (10) & 25.17$\pm$7.63 (7) & 14.45$\pm$2.53 (2) & 46.26$\pm$3.97 (9) & 47.58$\pm$3.27 (8) & 32.34$\pm$3.25 (9) & 7.9        \\

           \hline
          { \textbf{ Bagging$_{30}$} }    & 29.16$\pm$1.02 (7) & 28.58$\pm$2.81(6) & 46.59$\pm$1.39  (8)  & 21.72$\pm$4.24  (2) & 14.00$\pm$1.96 (1) & 45.85$\pm$2.23 (7)& 48.08$\pm$4.21 (10) & 27.86$\pm$2.27 (7)& 6.0 \\

		   \hline
           \textbf{  DRM$_{30}$}          & 26.05$\pm$1.14 (4) & 24.47$\pm$0.89 (2) & 14.43$\pm$1.65 (5)  & 26.55$\pm$3.91 (8) & 35.98$\pm$2.84  (10)& 36.67$\pm$4.03 (2) & 19.00$\pm$2.51 (6)  & 19.70$\pm$3.48 (6)& 5.4    \\
		  
           \hline
        \end{tabular}
        }
	 \end{center}	
	  \vspace{-10pt}
\end{table*}

\begin{table*}[t]
\caption{Average training time in seconds. }
\vspace{-10pt}
	\label{tab:time}
	\begin{center}
	\resizebox{\textwidth}{!}{
		\begin{tabular}{ c c c c c c c c c c}
		   \hline
		    { \textbf{L$_1$-ERM$_{10}$}} &  { \textbf{ L$_2$-ERM$_{10}$}}  & {\textbf{AdaBoost$_{10}$} } & {\textbf{ Bagging$_{10}$} }  &  { \textbf{DRM$_{10}$ }  } &  { \textbf{L$_1$-ERM$_{30}$}} &  { \textbf{ L$_2$-ERM$_{30}$}}  & {\textbf{AdaBoost$_{30}$} } & {\textbf{ Bagging$_{30}$} } &  { \textbf{DRM$_{30}$ }  } \\
		   \hline
		  0.0568 & 0.0575 & 0.1758 & 0.3019 &168.5657 & 0.1358 & 0.1091 & 0.2841 & 0.2512 & 193.9761\\
           \hline
        \end{tabular}
        }
	 \end{center}	
	 \vspace{-20pt}	
\end{table*}

%\hline
%           L$_1$-PSVM             & 26.87$\pm$1.33 & 24.56$\pm$1.01 & 14.41$\pm$0.71 & 26.03$\pm$5.72 & 28.35$\pm$2.42 & 39.28$\pm$4.43 & 18.33$\pm$2.39  & 13.13$\pm$2.15     & 1 \\
%		   \hline
%           L$_2$-PSVM             & 30.60$\pm$1.78 & 24.81$\pm$1.11 & 15.31$\pm$2.59 & 32.93$\pm$8.39 & 36.18$\pm$4.24 & 35.13$\pm$3.25 & 18.67$\pm$2.67 & 15.32$\pm$2.44 &2         \\
%           \hline
%          $\nu$-SVM                  & 26.96$\pm$1.01 & 24.69$\pm$0.80 & 14.07$\pm$1.08 & 26.38$\pm$2.31 & 34.91$\pm$2.84 & 36.15$\pm$3.82 & 18.92$\pm$1.36 & 19.85$\pm$3.24 &3          \\

\textbf{Performance Comparison }
 This part compares our proposed ERM with the classic ensemble models including AdaBoost and Bagging, and the recently designed DRM. {The codes of DRM are downloaded from the authors' website, while those of AdaBoost and Bagging are integrated in the Matlab statistics toolbox (\textit{fitensemble} function). The base of DRM, $\nu$-SVM, is from LibSVM.}  The sizes and distributions of the datasets are various, to avoid the effect brought by the amount of training data and test the generalization ability of the ensembles learned from different types of data, the number of training samples for all the datasets is fixed to $150$.

Table \ref{tab:acc} provides the quantitative comparison among the competitors. We report the mean testing errors over $10$ independent trials by randomly sampling $150$ data points from a dataset as its training set and the rest as the testing. AdaBoost and Bagging are inferior to ERM and DRM in most cases. The exception is on the \textit{splice} dataset. As for our ERM, we can see that it significantly outperforms the others on the \textit{australian}, \textit{sonar}, \textit{heart} and \textit{ionosphere}, and competes very favorably on the \textit{german} dataset. On each dataset, we assign ranks to methods. And the average ranks (A.R.) of the competitors over the involved datasets are given in the last column of Tab. \ref{tab:acc}. It can be observed that the top five average ranks are all lower than $5.0$, and four of which are from ERM methods. The best and the second best belong to L$_2$-ERM$_{30}$ (A.R.=2.5) and L$_2$-ERM$_{10}$ (A.R.=3.4) respectively, while the fourth and fifth places are taken by L$_1$-ERM$_{10}$ (A.R.=4.8) and L$_1$-ERM$_{30}$ (A.R.=4.9) respectively. The third goes to DRM$_{10}$, the average rank of which is $4.6$. The results on the \textit{ijcnn1} are not included in the table, because all the competitors perform very closely to each other, which may lead to an unreliable rank. 

Another issue should be concerned is the efficiency. Table \ref{tab:time} lists the mean training time over all the datasets and each dataset executes $10$ runs. From the numbers, we can see the clear advantage of our ERM. L$_1$-ERM$_{10}$ and L$_2$-ERM$_{10}$ only spend about $0.05s$ on training, while the ERMs with $30$ components, \textit{i.e.} L$_1$-ERM$_{30}$ and L$_2$-ERM$_{30}$, cost less than $0.14s$. Both AdaBoost and Bagging are sufficiently efficient, which take less than $0.3s$ to accomplish the task. But the training uses $168.57s$ and $193.97s$ by DRM for the $10$-base and $30$-base cases respectively, which are about $2000$ times expensive as the proposed ERM. We would like to mention that the core of DRM is implemented in C++, while our ERM is in pure Matlab. Moreover, as theoretically analyzed in Sec. \ref{sec:ta} and empirically verified in Sec. \ref{sec:c&t}, our algorithm is (quasi) linear with respect to the size of training set. In other words, the merit of ERM in time would become more conspicuous as the scale of training data increases, in comparison with AdaBoost, Bagging and DRM. {Due to space limit, only several experiments are shown in the paper to demonstrate the efficacy of our ERM. To allow more experimental verification, our code can be downloaded from \url{http://cs.tju.edu.cn/orgs/vision/~xguo/homepage.htm} }

\section{Conclusion}
\label{sec:c&d}
The diversity of component learners is critical to the ensemble performance. This paper has defined a new measurement of diversity, \textit{i.e.} exclusivity. Incorporating the designed regularizer with the hinge loss function gives a birth to a novel model, namely Exclusivity Regularized Machine. The convergence of the proposed ALM-based algorithm to a global optimal solution is theoretically guaranteed. The experimental results on several benchmark datesets, compared to the state-of-the-arts, have demonstrated the clear advantages of our method in terms of accuracy and efficiency.

Our framework is ready to embrace more elaborate treatments for further improvement. For instance, due to the relationship $\|\bs{u}\|_1=\mc{X}_r(\bs{u},\bs{1})$, as discussed in Sec. \ref{sec:d&f}, the sparsity on $\bs{W}$ can be promoted by extending $\tilde{\bs{W}}$ to $[\bs{W},\beta\bs{1}]$, where $\beta$ is a weight coefficient of the sparsity. In addition, it is difficult to directly solve the $\bs{E}$ sub-problem \eqref{eq:E} with arbitrary given $p$. Fortunately, in this work, it is always that $p\geq 1$. Thus the partial derivative of \eqref{eq:E} with respect to $\bs{E}$ is monotonically increasing. The binary search method can be employed to narrow the possible range of $\bs{E}$ by half via each operation. It is positive that our ERM can be widely applied to various classification tasks. Although, for avoiding distractions, no experiments are provided to evaluate the performance of the possible variants, it is positive that our ERM can be widely applied to various classification tasks.

%, and obtain an $\epsilon-$accurate solution in $o(\log \frac{1}{\epsilon})$ time
\section*{Acknowledgment}
Xiaojie Guo would like to thank Dr. Ju Sun with Department of Electrical Engineering, Columbia University, for his suggestions on this work.

\small{
\bibliographystyle{ieeetr}
\bibliography{nips16.bib}
}

\end{document}